\documentclass[runningheads]{llncs}
\usepackage{graphicx}
\usepackage{amsfonts}
\usepackage{cite}
\usepackage{tabularx}
\usepackage{hyperref}

\usepackage{amsmath}
\DeclareMathOperator*{\argmax}{argmax}
\DeclareMathOperator*{\argmin}{argmin}

\begin{document}
\title{MaNi: Maximizing Mutual Information for Nuclei Cross-Domain Unsupervised Segmentation}
\titlerunning{MaNi}
%
\author{Yash Sharma\inst{1} \and 
Sana Syed\inst{1}\thanks{Co-Corresponding Author}\and 
Donald E. Brown\inst{1}\protect\footnotemark[1]} 
\authorrunning{Y. Sharma et al.}
\institute{University of Virginia, Charlottesville, Virginia, USA \\
\email{ys4yh@virginia.edu}\\
}

%
%

%
\maketitle              
\begin{abstract}

In this work, we propose a mutual information (MI) based unsupervised domain adaptation (UDA) method for the cross-domain nuclei segmentation. Nuclei vary substantially in structure and appearances across different cancer types, leading to a drop in performance of deep learning models when trained on one cancer type and tested on another. This domain shift becomes even more critical as accurate segmentation and quantification of nuclei is an essential histopathology task for the diagnosis/ prognosis of patients and annotating nuclei at the pixel level for new cancer types demands extensive effort by medical experts. To address this problem, we maximize the MI between labeled source cancer type data and unlabeled target cancer type data for transferring nuclei segmentation knowledge across domains. We use the Jensen-Shanon divergence bound, requiring only one negative pair per positive pair for MI maximization. We evaluate our set-up for multiple modeling frameworks and on different datasets comprising of over 20 cancer-type domain shifts and demonstrate competitive performance. All the recently proposed approaches consist of multiple components for improving the domain adaptation, whereas our proposed module is light and can be easily incorporated into other methods (Implementation: \href{https://github.com/YashSharma/MaNi}{https://github.com/YashSharma/MaNi}). 

\keywords{Unsupervised Domain Adaptation \and Contrastive Learning \and Histology \and Instance Segmentation \and Semantic Segmentation}
\end{abstract}

\section{Introduction}

Nuclei are the fundamental organizational unit of life. Accurate assessment of these nuclei in histopathology slides provides critical morphological information necessary for the quantitative analysis of multiple diseases. In the last decade, Convolutional Neural Network-based approaches have emerged as a promising solution in medical imaging tasks, and nuclei segmentation tasks \cite{graham2019hover, ronneberger2015u}. However, they require a large amount of annotated data for any reliable training. 
Moreover, even with a large dataset, models are prone to be poor at generalizing learned knowledge from one cancer type to another. Pixel-level image annotations by a medical expert are required to expand to a new cancer type, making it difficult, expensive, and a time-consuming task. Therefore, there is a need for methods that can transfer learned information from existing cancer datasets to other domains without additional annotations. We tackle this problem, characterized as Unsupervised domain adaptation (UDA), to close the gap between the annotated source domain and unlabeled target domain by learning domain-invariant and task-relevant features. 

It has been observed that trivially training a model on the source domain and evaluating on a target domain leads to sub-par performance owing to domain shift \cite{guan2021domain}. Moreover, the domain shift problem is widespread in medical image datasets due to different scanners, scanning protocols, and tissue types, among others. Hence, multiple strategies comprising of adversarial learning, pseudo-labeling, consistency regularization, or data-augmentation have been proposed to tackle this limitation \cite{guan2021domain}. Among these, self-training has emerged as a competitive approach. Various de-noising strategies such as confidence thresholding, uncertainty estimation, or distance penalty between domains have been proposed. These strategies have led to gains, but they all rely on unlabeled data points closer to labeled data points for pseudo-label training since those are the points with high confidence and attempts to push the target distribution closer to the source. However, forcing the target-domain distribution towards the source-domain distribution can destroy the latent structural patterns of the target domain, leading to a drop in performance. We hypothesize that instead of using these approaches, using an information-theoretic distance can lead to better alignment between the domains.


Contrastive learning (CL) has seen wide success in representation learning with applications ranging from unsupervised pre-training to multi-modal alignment \cite{shrivastava2021clip}. The basic idea of CL is to push together the latent distribution of similar samples and push away the latent distribution of dissimilar samples. 
Wang et al. \cite{wang2021exploring} adopted a supervised, pixel-wise contrastive learning algorithm and treated pixels belonging to the same class as positive pairs and pixels from dissimilar classes as negative pairs, observing significant gains in semantic segmentation performance. We take inspiration from their work and expand the idea of contrasting similar class pixels against dissimilar class pixels for UDA. 

In summary, our paper makes the following contributions:
1) We propose a simple Jensen-Shannon Divergence-based contrastive loss for UDA in nuclei semantic and instance segmentation tasks. The proposed loss maximizes the mutual information between the representations of ground truth nuclei pixels from the labeled source dataset and the pseudo-labeled nuclei pixels in the target data.
2) We demonstrate our approach using different architectures and for over 20 cancer-type domain shifts establishing that the inclusion of the MI loss leads to competitive gain over recently proposed methods.

\section{Related Works}

\subsection{Unsupervised Domain Adaptation}

UDA is a well-studied problem in literature where two of the widely adopted techniques are adversarial learning and self-training. In Adversarial learning, researchers attempt to align source data representation with target data representation via discriminator training. Hoffman et al. \cite{hoffman2018cycada} proposed a cycle-consistent adversarial domain adaptation (DA) method for enforcing cycle-consistency between domains. \cite{luo2019taking} used a category-level adversarial network for enforcing semantic consistency between each class. Tsai et al. \cite{tsai2019domain} developed a patch alignment method for DA by adversarially pushing the feature representations of clustered patches together. Vu et al. \cite{vu2019advent} used an entropy-based adversarial training approach for aligning weighted self-information distributions of different domains. Yang et al. \cite{yang2020adversarial} showed improvement in domain alignment by iteratively defensing against pointwise adversarial perturbations for domains. 


In Self-training, pseudo-labels are iteratively generated on target datasets using the model trained on labeled source datasets and used for retraining. 
Recent papers have incorporated different de-noising strategies for improving the accuracy of pseudo-labels.
Zhang et al. \cite{zhang2019category} used category centroids from the source domain for pseudo-labeling the target data and used the distance to centroids for training. Zhang et al. \cite{zhang2021prototypical} denoised pseudo-label by online correcting them according to the relative feature distance to the prototypes. Zou et al. \cite{zou2018unsupervised} used class-wise normalized confidence scores for generating pseudo-labels with balanced class distribution for self-training. Zou et al. \cite{zou2019confidence} proposed a confidence regularization for smoothing the prediction via regularizer loss minimization. 



We highlight the relevant works extending the above approaches for UDA in nuclei segmentation. Yang et al. \cite{yang2021minimizing} used adversarial domain discriminator and cyclic adaptation with pseudo labels for UDA. Further, they utilized weak labels to improve nuclei instance segmentation and classification. 
Haq et al. \cite{haq2020adversarial} used adversarial learning loss and reconstruction loss on output space for UDA for cell segmentation. Li et al. \cite{li2021unsupervised} extended \cite{haq2020adversarial} and applied a self-ensembling method with a student-teacher framework for imposing consistency loss along with reconstruction and adversarial training. 
In another branch of works, \cite{liu2020pdam} performed UDA from microscopy to histopathology images by synthesizing target-type images using CycleGAN followed by an inpainting module before applying adversarial adaptation. However, this approach requires pixel-level translation for synthesizing target-like images before a segmentation module and uses instance-level information for the mutual information-based feature alignment, constraining its adoption to only instance segmentation. Therefore, we limit our work and first test it extensively for tissue/ cancer type domain shifts and leave microscopy to histopathology adaptation for future works. Further, our proposed MI-based feature alignment strategy can be used for both semantic and instance segmentation problems. 


\subsection{Contrastive Learning and Mutual Information}

Contrastive learning (CL) methods are being widely adopted for classification and segmentation tasks of different modalities. 
For semi-supervised segmentation, Alonso et al. \cite{alonso2021semi} used positive-only CL for enforcing the segmentation network to yield similar pixel-level feature representation for same-class samples between labeled and unlabeled datasets. Hu et al. \cite{hu2021semi} for MRI and CT image segmentation pretraining used global contrastive loss on unlabeled images and supervised local contrastive loss on limited labeled images. For volumetric medical segmentation, Chaitanya et al. \cite{chaitanya2020contrastive} leveraged similarity across corresponding slices in different volumes for defining positive and negative pairs for contrastive loss. They used infoNCE bound for CL, requiring a large number of negative samples for training. For tackling this dependency, Peng et al. \cite{peng2021boosting} maximized mutual information (MI) on categorical distribution by projecting the continuous feature embedding to clustering space. They used encoder representation for global regularization and, for local regularization, maximized MI between neighboring feature vectors at multiple intermediate levels. 

Our work maximizes the MI between similar classes using labeled source images and pseudo-labeled target images. 
Concurrent to our work, Chaitanya et al. \cite{chaitanya2021local} proposed an end-to-end semi-segmentation framework by defining a local pixel-level contrastive loss between pseudo-labels of unlabeled sets and limited labeled sets. They randomly sample pixels from each image to address the computational limitations of running CL for all the pixels. In contrast, we use average pooling at the class level for considering all the pixels of labeled and unlabeled images. Further, we use JSD bound instead of InfoNCE for MI estimation and focus on domain adaptation tasks. Shrivastava et al. \cite{shrivastava2021clip} in their work for visual representation learning from textual annotation, demonstrated that the JSD-based bound enables the MI maximization at a smaller batch size with just one negative sample. Also, \cite{hjelm2018learning} demonstrated in their extended analysis that JSD is insensitive to the number of negative samples, while infoNCE shows a decline as the number of negative samples decreases, motivating us to choose JSD bound for MI estimation. 


\section{Methods}

\subsection{Problem Set-Up}

In our work, we tackle the problem of unsupervised domain adaptation for nuclei segmentation, where we have labeled source domain data and unlabeled target domain data coming from different cancer types. Our labeled source data has $N_s$ images $(x_s, y_s)$ and our unlabeled target domain has $N_t$ images $(x_t)$.

\subsection{Segmentation Loss and Mutual Information Maximization}




We use a combination of dice loss and binary cross-entropy loss for supervised segmentation training $(L_{seg})$. For mutual information (MI) maximization, we use Jensen-Shannon Divergence (JSD)-based lower bound proposed in \cite{hjelm2018learning}. This bound allows us to estimate the MI with just one negative example for each positive example. In our framework, we define nuclei pixels in the source domain with nuclei pixels in the target domain as positive pairs while background pixels in the source domain with nuclei pixels in the target domain as a negative pair. 

We use a simple $1\times1$ convolution network with the same number of input and output channels, followed by batch normalization and ReLU as the projection head. The output of the backbone network is passed to the projection head for generating representation for MI loss. Projected feature representation and segmentation labels are used for MI maximization. Since labels are not available for the target domain, we use a segmentation network output from the same iteration as pseudo-labels. Typically, we would like to contrast all the positive pair pixels with negative pair pixels. However, for computational feasibility, we mean pool the projected representations using the segmentation labels to get aggregated background and nuclei pixel representations. Further, mean pooled source nuclei representation with mean pooled target nuclei representation is treated as a positive pair, whereas mean pooled background pixel representation with mean pooled target nuclei representation is treated as a negative pair. We define our JSD estimator as: 

\begin{equation}
\hat{I}_{\omega}^{JSD}(Z_s; Z_t) := \mathbb{E}_{P(Z_s,Z_t)}[-sp(-T_{\omega}(z_s, z_t))] - \mathbb{E}_{P(Z_s)P(Z_t)}[sp(T_{\omega}(z_s', z_t))]
\end{equation}

where $z_s$ is mean pooled source nuclei representation, $z_t$ is mean pooled target nuclei representation, $z_s'$ is mean pooled background representation for the source sample, and $sp(z) = log(1+e^z)$ is the softplus function. Here, $T_{\omega}: Z_s \times Z_t \to \mathbb{R} $ is a discriminator network with trainable parameters $\omega$ which are jointly optimized to distinguish between a paired-sample from a joint distribution (positive pair) and a paired sample from product of marginals (negative pair). For discriminator network, we use the concat architecture proposed in \cite{hjelm2018learning}.  

\begin{figure}[t]
\includegraphics[width=\textwidth]{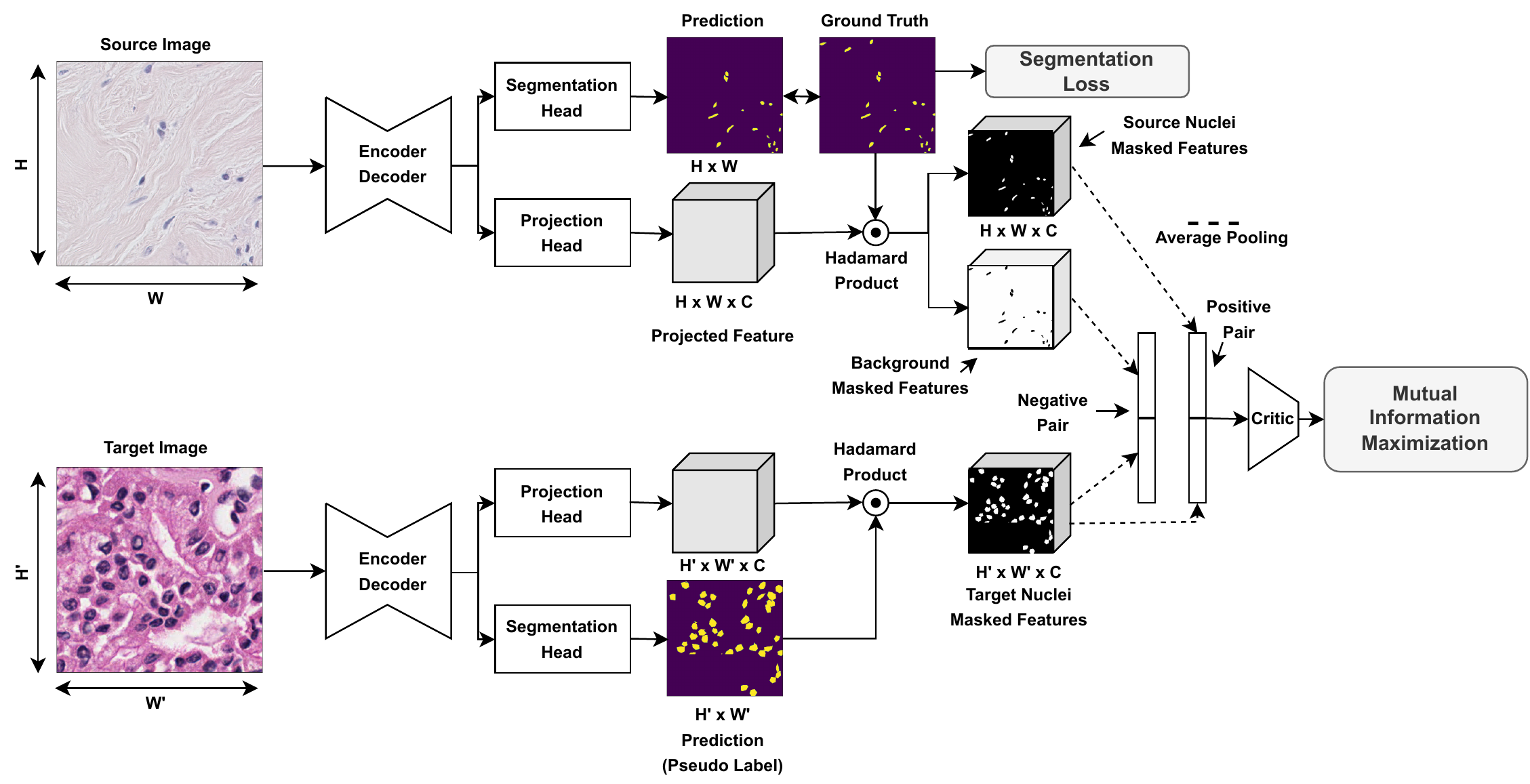}
\caption{Images pass through the backbone network and segmentation head for the segmentation training and projection head for contrastive training. Source image actual label is used for segmentation loss and to obtain the feature representation of nuclei and background area. Target image prediction is used as the pseudo label to get the feature representation of the nuclei area. Masked features are average pooled for generating positive and negative pairs for mutual information maximization.} \label{fig1}
\end{figure}

\subsection{Training}

As shown in Figure 1, our architecture is divided into 4 parts - 1) backbone encoder-decoder network ($\theta_b$), 2) segmentation head ($\theta_s$) for generating segmentation masks, 3) projection head ($\theta_p$) for generating feature representations for MI maximization and 4) discriminator network $(T_\omega)$ for estimating MI. 

We divide our training into 2 steps. In step 1, for warming up the model and achieving reasonable pseudo-labels for target data, we pretrain the backbone network ($\theta_b$) and segmentation head ($\theta_s$) using source data on segmentation loss.

\begin{equation}
(\hat{\theta_b}, \hat{\theta_s}) = \argmin_{\theta_b, \theta_s}\frac{1}{\left| X_s \right|} \sum_{(x_s, y_s)}^{} L_{seg} (y_s, \hat{y_s})
\end{equation}

In step 2, we continue the training the model with the MI loss between labeled source data points and pseudo-labeled target data points. 

\begin{equation}
    \begin{split}
    (\hat{\theta_b}, \hat{\theta_s}, \hat{\theta_p}, \omega) = \argmin_{\theta_b, \theta_s}\frac{1}{\left| X_s \right|} \sum_{(x_s, y_s)}^{} L_{seg} (y_s, \hat{y_s}) +\\ \argmax_{\omega, \theta_p, \theta_b, \theta_s} \frac{1}{\left| X_{pair} \right|} \sum_{(x_s, y_s), (x_t, \hat{y_t})}^{} \hat{I}_{\omega}^{JSD}(Z_s; Z_t) 
    \end{split}
\end{equation}

where $Z_s$ and $Z_t$ are mean pooled nuclei representation for source and target respectively, and $X_{pair}$ define randomly paired source and target image for MI maximization. 

\section{Experiments}

\subsection{Dataset and Implementation}

We evaluate our approach using the similar setting used in \cite{li2021unsupervised} for Nuclei Semantic Segmentation and \cite{yang2021minimizing} for Nuclei Instance Segmentation. In \cite{li2021unsupervised}, authors used the Kidney Renal Clear Cell Carcinoma (KIRC) dataset \cite{irshad2014crowdsourcing}, Triple Negative Breast Cancer Cell (TNBC) dataset \cite{naylor2018segmentation}, and Stomach adenocarcinoma (STAD) \cite{hou2020dataset} of the TCIA repository. There are $486$ images of $400\times400$ pixel size in KIRC, $50$ images of $512\times512$ pixel size in TNBC, and $99$ images of $256\times256$ pixel size in STAD. \cite{li2021unsupervised} evaluated UDA Nuclei Semantic Segmentation for following cancer type domain shift - TNBC to KIRC/ TCIA, and TCIA to KIRC/ TNBC. Following the same setup for comparison, we use standard U-Net architecture as the backbone encoder-decoder network and $80\%$ of images for training, $10\%$ for validation, and $10\%$ for testing. We report average dice scores from five runs with different splits for accounting for sampling bias.
 
In \cite{yang2021minimizing}, the authors used colorectal nuclear segmentation and phenotype dataset (CoNSep) \cite{graham2019hover} as the source domain and PanNuke dataset \cite{gamper2019pannuke} comprising 19 cancer types as the target domain for UDA in nuclei instance segmentation and classification task. 
As used in their paper, we use labeled CoNSep and the first split of PanNuke for training, second split for validation, and third split for testing. We focused on the instance segmentation task and didn't use classification labels for training. We use HoverNet as the backbone encoder-decoder network with MI maximization module included in the nuclei segmentation branch. We report Dice Score, Aggregate Jaccard Index (AJI), Panoptic Quality (PQ), Detection Quality (DQ), and Segmentation Quality (SQ) for comparison.

All experiments are carried out in PyTorch using $1$ A100 GPU for nuclei semantic segmentation and $4$ A100 GPUs for nuclei instance segmentation. For the nuclei semantic segmentation experiment, we train the model for $1000$ iterations using only segmentation loss and $9000$ iterations with both segmentation and MI maximization loss. We use a batch size of $4$, Adam optimizer with a learning rate of $1e-3$, and rotation augmentation. We randomly pair labeled source images with unlabeled target images during the MI maximization step. For nuclei instance segmentation experiment, we follow the training details provided in \cite{yang2021minimizing} and \cite{graham2019hover}. We extend our approach only for nuclei instance segmentation and will focus on classification in our future work. First, we freeze the encoder and train the decoder for $50$ epochs, with the first $10$ epochs only using segmentation loss and the subsequent $40$ epochs using both segmentation and MI loss. Then we unfreeze and train the whole network for $50$ epochs with both losses. We use a batch size of $16$ and Adam optimizer with a learning rate of $1e-4$.

\subsection{Evaluation}

\begin{table}[t]
\caption{Nuclei Semantic Segmentation Results for UDA.}\label{tab1}
\centering
\begin{tabular}{|c|c|c|c|c|}
\hline
\textbf{Source Domain} & \multicolumn{2}{c|}{\textbf{TNBC}} & \multicolumn{2}{c|}{\textbf{TCIA}}\\
\hline
\textbf{Target Domain} & \textbf{KIRC} & \textbf{TCIA} & \textbf{KIRC} & \textbf{TNBC}\\
\hline
Source Only & 0.713 & 0.680 & 0.710 & 0.791\\
DA\_ADV \cite{dong2018unsupervised} & 0.726 & 0.734 & 0.708 & 0.787\\
CellSegUDA \cite{haq2020adversarial} & 0.728 & 0.765 & 0.705 & 0.805\\
SelfEnsemblingUDA \cite{li2021unsupervised} & 0.727 & 0.761 & 0.715 & 0.816\\
MaNi & 0.733 & 0.776 & 0.727 & 0.821\\
\hline
\end{tabular}
\end{table}

\subsubsection{Nuclei Semantic Segmentation} - Table~\ref{tab1} compares our approach with recently proposed approaches for UDA for nuclei semantic segmentation \footnote{Other approach results reported using best self-implementation}. MaNi outperforms all other approaches for both TNBC to KIRC/ TCIA shift and TCIA to KIRC/ TNBC shift. The next best performing approach proposed in \cite{li2021unsupervised} uses multiple components consisting of a discriminator and a reconstruction network, consistency regularization between perturbed target images, and post-processing with CRF. While MaNi only uses the MI maximization module with segmentation loss for training, reducing the complexity of the architecture. We attribute high gains to the MI module's ability to transfer nuclei segmentation knowledge from labeled source pixels to pseudo-labeled target pixels. 

\begin{table}[t]
\caption{Nuclei Instance Segmentation Results for UDA from CoNSep to PanNuke.}\label{tab2}
\centering
\begin{tabular}{|c|c|c|c|c|c|}
\hline
\textbf{Method} & \textbf{Dice} & \textbf{AJI} & \textbf{DQ} & \textbf{SQ} & \textbf{PQ} \\
\hline
Source Only & 0.576 & 0.387 & 0.461 & 0.657 & 0.342\\
GRL \cite{ganin2015unsupervised} & 0.723 & 0.509 & 0.587 & 0.756 & 0.450\\
Paul et al. \cite{paul2020domain} & 0.731 & 0.501 & 0.600 & 0.751 & 0.446\\
Yang et al. \cite{yang2021minimizing} & 0.740 & 0.516 & 0.602 & 0.753 & 0.460\\
MaNi & 0.735 & 0.534 & 0.621 & 0.742 & 0.477\\
\hline
\end{tabular}
\end{table}

\subsubsection{Nuclei Instance Segmentation} - Results for UDA in nuclei instance segmentation from CoNSep data to PanNuke data is reported in Table~\ref{tab2}\footnote{Other approach results reported from \cite{yang2021minimizing}}. MaNi performs competitively to the best performing approach. MaNi demonstrates superior performance over PQ, DQ, and AJI, highlighting that our approach can detect and segment more nuclei accurately and suffers less from the issue of false negatives. While in terms of SQ and Dice Score, MaNi is competitive to other approaches. 

\begin{table}
\parbox{.45\linewidth}{
\centering
\caption{Impact of MI loss weight.}\label{tab3}
\begin{tabular}{|c|c|c|}
\hline
\textbf{Source Domain} & \textbf{TNBC} & \textbf{TCIA} \\
\hline
\textbf{Target Domain} & \textbf{TCIA} & \textbf{TNBC} \\
\hline
1 & 0.776 & 0.821 \\
0.1 & 0.769 & 0.765 \\
0.01 & 0.751 & 0.765 \\
\hline
\end{tabular}
}
\hfill
\parbox{.45\linewidth}{
\centering
\caption{Impact of type of MI loss.}\label{tab4}
\begin{tabular}{|c|c|c|}
\hline
\textbf{Source Domain} & \textbf{TNBC} & \textbf{TCIA} \\
\hline
\textbf{Target Domain} & \textbf{TCIA} & \textbf{TNBC} \\
\hline
Mean-Pooling & 0.776 & 0.821 \\
Max-Pooling & 0.751 & 0.817 \\
Random Pixels & 0.771 & 0.809 \\
\hline
\end{tabular}
}
\end{table}

\subsection{Ablation Studies}

We study the impact of different weightage on the MI loss for training. As shown in Table $3$, the default choice of $1$ performs best. The impact of MI loss dilutes with reduction in weight, leading to a drop in performance. Further, we study how replacing the average pooling approach with the max-pooling approach or sampling random pixels approach (N=$4$), as done in \cite{chaitanya2021local}, impacts performance. Average pooling performs superior to other approaches.

\section{Conclusion}

In this paper, we propose a JSD-based MI loss for UDA for Nuclei Semantic Segmentation and Instance Segmentation. We perform comprehensive experiments with different architecture - UNet and HoverNet and for different cancer-type domain shifts - TNBC to TCIA/KIRC, TCIA to KIRC/TNBC, and CoNSep to PanNuke. We highlight that our method leads to gain for both semantic segmentation and instance segmentation. We plan to extend this approach for nuclei classification tasks and subsequently to general imaging tasks in our future work.

\subsubsection{Acknowledgements} This work was supported by NIDDK of the National Institutes of Health under award number K23DK117061-01A1 and Litwin IBD Pioneers Award of the Crohn’s \& Colitis Foundation. 

%
%
%
\bibliographystyle{splncs04}
\bibliography{paper2233}

\begin{thebibliography}{10}
\providecommand{\url}[1]{\texttt{#1}}
\providecommand{\urlprefix}{URL }
\providecommand{\doi}[1]{https://doi.org/#1}

\bibitem{alonso2021semi}
Alonso, I., Sabater, A., Ferstl, D., Montesano, L., Murillo, A.C.:
  Semi-supervised semantic segmentation with pixel-level contrastive learning
  from a class-wise memory bank. In: Proceedings of the IEEE/CVF International
  Conference on Computer Vision. pp. 8219--8228 (2021)

\bibitem{chaitanya2020contrastive}
Chaitanya, K., Erdil, E., Karani, N., Konukoglu, E.: Contrastive learning of
  global and local features for medical image segmentation with limited
  annotations. Advances in Neural Information Processing Systems  \textbf{33},
  12546--12558 (2020)

\bibitem{chaitanya2021local}
Chaitanya, K., Erdil, E., Karani, N., Konukoglu, E.: Local contrastive loss
  with pseudo-label based self-training for semi-supervised medical image
  segmentation. arXiv preprint arXiv:2112.09645  (2021)

\bibitem{dong2018unsupervised}
Dong, N., Kampffmeyer, M., Liang, X., Wang, Z., Dai, W., Xing, E.: Unsupervised
  domain adaptation for automatic estimation of cardiothoracic ratio. In:
  International conference on medical image computing and computer-assisted
  intervention. pp. 544--552. Springer (2018)

\bibitem{gamper2019pannuke}
Gamper, J., Koohbanani, N.A., Benet, K., Khuram, A., Rajpoot, N.: Pannuke: an
  open pan-cancer histology dataset for nuclei instance segmentation and
  classification. In: European Congress on Digital Pathology. pp. 11--19.
  Springer (2019)

\bibitem{ganin2015unsupervised}
Ganin, Y., Lempitsky, V.: Unsupervised domain adaptation by backpropagation.
  In: International conference on machine learning. pp. 1180--1189. PMLR (2015)

\bibitem{graham2019hover}
Graham, S., Vu, Q.D., Raza, S.E.A., Azam, A., Tsang, Y.W., Kwak, J.T., Rajpoot,
  N.: Hover-net: Simultaneous segmentation and classification of nuclei in
  multi-tissue histology images. Medical Image Analysis  \textbf{58},  101563
  (2019)

\bibitem{guan2021domain}
Guan, H., Liu, M.: Domain adaptation for medical image analysis: a survey. IEEE
  Transactions on Biomedical Engineering  (2021)

\bibitem{haq2020adversarial}
Haq, M.M., Huang, J.: Adversarial domain adaptation for cell segmentation. In:
  Medical Imaging with Deep Learning. pp. 277--287. PMLR (2020)

\bibitem{hjelm2018learning}
Hjelm, R.D., Fedorov, A., Lavoie-Marchildon, S., Grewal, K., Bachman, P.,
  Trischler, A., Bengio, Y.: Learning deep representations by mutual
  information estimation and maximization. arXiv preprint arXiv:1808.06670
  (2018)

\bibitem{hoffman2018cycada}
Hoffman, J., Tzeng, E., Park, T., Zhu, J.Y., Isola, P., Saenko, K., Efros, A.,
  Darrell, T.: Cycada: Cycle-consistent adversarial domain adaptation. In:
  International conference on machine learning. pp. 1989--1998. PMLR (2018)

\bibitem{hou2020dataset}
Hou, L., Gupta, R., Van~Arnam, J.S., Zhang, Y., Sivalenka, K., Samaras, D.,
  Kurc, T.M., Saltz, J.H.: Dataset of segmented nuclei in hematoxylin and eosin
  stained histopathology images of ten cancer types. Scientific data
  \textbf{7}(1),  1--12 (2020)

\bibitem{hu2021semi}
Hu, X., Zeng, D., Xu, X., Shi, Y.: Semi-supervised contrastive learning for
  label-efficient medical image segmentation. In: International Conference on
  Medical Image Computing and Computer-Assisted Intervention. pp. 481--490.
  Springer (2021)

\bibitem{irshad2014crowdsourcing}
Irshad, H., Montaser-Kouhsari, L., Waltz, G., Bucur, O., Nowak, J., Dong, F.,
  Knoblauch, N.W., Beck, A.H.: Crowdsourcing image annotation for nucleus
  detection and segmentation in computational pathology. In: Pacific symposium
  on biocomputing Co-chairs. pp. 294--305. World Scientific (2014)

\bibitem{li2021unsupervised}
Li, C., Zhou, Y., Shi, T., Wu, Y., Yang, M., Li, Z.: Unsupervised domain
  adaptation for the histopathological cell segmentation through
  self-ensembling. In: MICCAI Workshop on Computational Pathology. pp.
  151--158. PMLR (2021)

\bibitem{liu2020pdam}
Liu, D., Zhang, D., Song, Y., Zhang, F., O’Donnell, L., Huang, H., Chen, M.,
  Cai, W.: Pdam: A panoptic-level feature alignment framework for unsupervised
  domain adaptive instance segmentation in microscopy images. IEEE Transactions
  on Medical Imaging  \textbf{40}(1),  154--165 (2020)

\bibitem{luo2019taking}
Luo, Y., Zheng, L., Guan, T., Yu, J., Yang, Y.: Taking a closer look at domain
  shift: Category-level adversaries for semantics consistent domain adaptation.
  In: Proceedings of the IEEE/CVF Conference on Computer Vision and Pattern
  Recognition. pp. 2507--2516 (2019)

\bibitem{naylor2018segmentation}
Naylor, P., La{\'e}, M., Reyal, F., Walter, T.: Segmentation of nuclei in
  histopathology images by deep regression of the distance map. IEEE
  transactions on medical imaging  \textbf{38}(2),  448--459 (2018)

\bibitem{paul2020domain}
Paul, S., Tsai, Y.H., Schulter, S., Roy-Chowdhury, A.K., Chandraker, M.: Domain
  adaptive semantic segmentation using weak labels. In: European conference on
  computer vision. pp. 571--587. Springer (2020)

\bibitem{peng2021boosting}
Peng, J., Pedersoli, M., Desrosiers, C.: Boosting semi-supervised image
  segmentation with global and local mutual information regularization. arXiv
  preprint arXiv:2103.04813  (2021)

\bibitem{ronneberger2015u}
Ronneberger, O., Fischer, P., Brox, T.: U-net: Convolutional networks for
  biomedical image segmentation. In: International Conference on Medical image
  computing and computer-assisted intervention. pp. 234--241. Springer (2015)

\bibitem{shrivastava2021clip}
Shrivastava, A., Selvaraju, R.R., Naik, N., Ordonez, V.: Clip-lite: Information
  efficient visual representation learning from textual annotations. arXiv
  preprint arXiv:2112.07133  (2021)

\bibitem{tsai2019domain}
Tsai, Y.H., Sohn, K., Schulter, S., Chandraker, M.: Domain adaptation for
  structured output via discriminative patch representations. In: Proceedings
  of the IEEE/CVF International Conference on Computer Vision. pp. 1456--1465
  (2019)

\bibitem{vu2019advent}
Vu, T.H., Jain, H., Bucher, M., Cord, M., P{\'e}rez, P.: Advent: Adversarial
  entropy minimization for domain adaptation in semantic segmentation. In:
  Proceedings of the IEEE/CVF Conference on Computer Vision and Pattern
  Recognition. pp. 2517--2526 (2019)

\bibitem{wang2021exploring}
Wang, W., Zhou, T., Yu, F., Dai, J., Konukoglu, E., Van~Gool, L.: Exploring
  cross-image pixel contrast for semantic segmentation. In: Proceedings of the
  IEEE/CVF International Conference on Computer Vision. pp. 7303--7313 (2021)

\bibitem{yang2020adversarial}
Yang, J., Xu, R., Li, R., Qi, X., Shen, X., Li, G., Lin, L.: An adversarial
  perturbation oriented domain adaptation approach for semantic segmentation.
  In: Proceedings of the AAAI Conference on Artificial Intelligence. vol.~34,
  pp. 12613--12620 (2020)

\bibitem{yang2021minimizing}
Yang, S., Zhang, J., Huang, J., Lovell, B.C., Han, X.: Minimizing labeling cost
  for nuclei instance segmentation and classification with cross-domain images
  and weak labels. In: Proceedings of the AAAI Conference on Artificial
  Intelligence. vol.~35, pp. 697--705 (2021)

\bibitem{zhang2021prototypical}
Zhang, P., Zhang, B., Zhang, T., Chen, D., Wang, Y., Wen, F.: Prototypical
  pseudo label denoising and target structure learning for domain adaptive
  semantic segmentation. In: Proceedings of the IEEE/CVF Conference on Computer
  Vision and Pattern Recognition. pp. 12414--12424 (2021)

\bibitem{zhang2019category}
Zhang, Q., Zhang, J., Liu, W., Tao, D.: Category anchor-guided unsupervised
  domain adaptation for semantic segmentation. Advances in Neural Information
  Processing Systems  \textbf{32} (2019)

\bibitem{zou2018unsupervised}
Zou, Y., Yu, Z., Kumar, B., Wang, J.: Unsupervised domain adaptation for
  semantic segmentation via class-balanced self-training. In: Proceedings of
  the European conference on computer vision (ECCV). pp. 289--305 (2018)

\bibitem{zou2019confidence}
Zou, Y., Yu, Z., Liu, X., Kumar, B., Wang, J.: Confidence regularized
  self-training. In: Proceedings of the IEEE/CVF International Conference on
  Computer Vision. pp. 5982--5991 (2019)

\end{thebibliography}
%




\end{document}


\title{Supplementary Material}
\titlerunning{MaNi}

\author{Anonymous}
\authorrunning{F. Author et al.}
%
\institute{Anonymous Organization \\
\email{***@*****.***}\\
}
\maketitle              
\section{First Section}